\documentclass[11pt,a4paper]{article}
\usepackage[hyperref]{acl2021}
\usepackage{times}
\usepackage{latexsym}

\usepackage{booktabs}
\usepackage{graphicx}
\usepackage{wtref}
\usepackage{amsmath,bm}
\usepackage{amsfonts}
\usepackage{etoolbox}
\usepackage{url}
\usepackage{algorithm, algpseudocode}
\usepackage{xcolor}
\usepackage{enumerate}
\usepackage{microtype}

\aclfinalcopy 

\usepackage{gb4e}
\noautomath

\newref{eq,fig,tab,sec}

\title{Effective Batching for Recurrent Neural Network Grammars}

\author{Hiroshi Noji \\
  Artificial Intelligence Research Center \\
  AIST \\
  \texttt{hiroshi.noji@aist.go.jp} \\\And
  Yohei Oseki \\
  Graduate School of Arts and Sciences \\
  University of Tokyo \\
  \texttt{oseki@g.ecc.u-tokyo.ac.jp} \\}

\date{}

\begin{document}
\maketitle

\begin{abstract}
As a language model that integrates traditional symbolic operations and flexible neural representations, recurrent neural network grammars (RNNGs) have attracted great attention from both scientific and engineering perspectives. However, RNNGs are known to be harder to scale due to the difficulty of batched training. In this paper, we propose effective batching for RNNGs, where every operation is computed in parallel with tensors across multiple sentences. Our PyTorch implementation effectively employs a GPU and achieves x6 speedup compared to the existing C++ DyNet implementation with model-independent auto-batching. Moreover, our batched RNNG also accelerates inference and achieves x20-150 speedup for beam search depending on beam sizes. Finally, we evaluate syntactic generalization performance of the scaled RNNG against the LSTM baseline, based on the large training data of 100M tokens from English Wikipedia and the broad-coverage targeted syntactic evaluation benchmark.\footnote{Our RNNG implementation is available at https://github.com/aistairc/rnng-pytorch/.}
\end{abstract}

\section{Introduction}
\seclabel{intro}
Neural language models have an excellent word prediction ability, which motivates researchers to develop several analysis methods for fine-grained evaluation, aiming at understanding which linguistic abilities the models have acquired during training~\cite{linzen-etal-2016-assessing,wilcox-etal-2018-rnn,marvin-linzen-2018-targeted,warstadt-etal-2020-blimp}.
So far, many efforts have been made on the evaluation of \textit{syntactic} performance of models, including the abilities to resolve distant subject-verb number agreement in English.
Since neural language models are the foundation of contemporary NLP systems, building a language model having robust sentence processing abilities like humans is an important goal, especially toward a system with human-like syntactic generalization abilities, not relying on the data-specific superficial cues found in the training data~\cite{mccoy-etal-2019-right,linzen-2020-accelerate}.

Past work has revealed that while sequential and unstructured models, such as LSTM and Transformer language models~\cite{hochreiter1997long,NIPS2017-3f5ee243}, can induce several interesting syntactic behaviors, there is also a notable advantage in explicitly modeling syntax with specific architectures~\cite{kuncoro-etal-2018-lstms,wilcox-etal-2019-structural,hu-etal-2020-systematic}.
The representative of such models is the recurrent neural network grammars (RNNGs;~\citealp{dyer-etal-2016-recurrent}), the top-down, left-to-right generative models of a parse tree and sentence.

While these results may suggest that RNNGs are a better modeling choice for language, unfortunately, they have a practical drawback in terms of scalability, due to their structure-sensitive computation mechanism~\cite{kuncoro-etal-2019-scalable}.
Since the computational graphs of RNNGs depend on the tree structures of the sentences, training cannot be mini-batched easily.
This is in contrast to LSTMs and Transformers, for which token-wise operations can be batched across sentences, allowing efficient computation on GPUs, which is the key to the data scalability.
Although RNNGs are claimed to be a fascinating language model, in practice, they still do not replace the unstructured, computationally favorable models like LSTMs.

In this paper, we directly address the data scalability issue of RNNGs by showing that most computations during training can be batched across sentences.
At  the computational core of RNNGs is stack LSTMs~\cite{dyer-etal-2015-transition}.
In past work, \citet{ding-koehn-2019-parallelizable} have already shown that stack LSTM update operations can be reduced to a tensor operation by implementing the stack as a single tensor with predefined maximum stack depth.
Our work is built on this idea but with a few additional techniques to bridge the gap between the simple stack LSTMs and RNNGs.
Importantly, we devise the efficient batching method for \textit{composition} operations on the arbitrary number of stack items, which is unsolved in previous work.

The existing RNNG implementation is based on DyNet that supports the mechanism called Autobatch~\cite{NIPS2017-c902b497}, which automatically finds mini-batch units from the independent computational graphs over multiple sentences with lazy computation.
While this mechanism is model-independent and allows intuitive implementation, the utility of this method rapidly plateaus as we increase the batch size.
On the other hand, our present method allows effective parallel computation, increasing the training speed almost linearly as we increase the batch size.

In addition to this new batching mechanism for improved scalability, we also provide a new analysis on the role of the strong syntactic inductive bias for models that can access the larger amount of data.
For syntactic generalization abilities, while \citet{hu-etal-2020-systematic} suggest that the model inductive bias plays a more important role than data scale, they also report that LSTMs or Transformers such as the off-the-shelf large-scale models (e.g., GPT-2 or JRNN) perform much better than their scale-controlled LSTMs.
Does an RNNG, which already works relatively well on a modest amount of data, still benefit from the data scale to further strengthen its syntactic ability?
We train a new RNNG on about 100M tokens in Wikipedia and evaluate its syntactic performance on SyntaxGym test circuits~\cite{gauthier-etal-2020-syntaxgym}, finding that the data scale generally brings further performance gains, while the model tends to lose some heuristics on surface patterns that LSTMs seem to find.
Our result suggests that RNNGs' reliance on structures will be strengthened with more data, motivating future research on developing better syntactic representation itself as supervision to structured language models.

A related approach to our work is adding the syntactic bias into \textit{sequential} language models, such as LSTMs, with knowledge distillation from RNNGs~\cite{kuncoro-etal-2019-scalable,doi:10.1162/tacl-a-00345}.
While motivations are similar, we provide a rather direct solution to resolve the scalability issue of RNNGs, opening up a new possibility of directly using them as an alternative to LSTMs.

From another perspective, our work can be complementary to this work, because knowledge distillation requires a teacher RNNG model, which itself is costly to obtain.
For example, \citet{doi:10.1162/tacl-a-00345} trained an RNNG on a relatively large dataset of 3.6M sentences, which is approximately similar to the training data we use.
While the detail is missing, they report that training takes three weeks on a GPU.
On the other hand, our models get almost converged in three days.
This direct improvement in training time greatly expands the applicability of RNNGs including a teacher of sequential models, and more direct use in computational psycholinguistics~\cite{hu-etal-2020-systematic} and NLP applications such as syntactic neural machine translation~\cite{eriguchi-etal-2017-learning}.

\section{Preliminaries}

\subsection{Recurrent neural network grammars}
\seclabel{rnng}
RNNGs are joint generative models of a sentence and constituency tree.
While RNN language models assign a next token probability, RNNGs assign a probability to next \textit{action}, by which the parse state (stack LSTM) changes dynamically.
In this work, we focus on the stack-only RNNG~\cite{kuncoro-etal-2017-recurrent}, which has some resemblance to RNNs in that a single state vector $\mathbf h_t$ defines next action probability $a_t$:
\begin{align*}
 a_t \sim \textrm{softmax}(\mathbf W_a \textrm{MLP}(\mathbf h_t) + \mathbf b_a)
\end{align*}

At each step, $\mathbf h_t$ is obtained from the top element of stack LSTM, which preserves intermediate LSTM states up to $\mathbf h_t$.
As a preparation for our batched RNNGs (Section~\secref{proposed}), we try to formalize how this stack LSTM states change with each action.
An RNNG internally preserves two different stacks: $S_h$ and $S_e$.
$S_h$ is a stack LSTM, keeping the LSTM hidden states $\mathbf h_0 \cdots \mathbf h_t$.\footnote{
Precisely, we also have to keep LSTM cell states.
We omit this part for brevity.
}
$S_e$ keeps stack elements, each of which is a word embedding $\mathbf e_w$, an open nonterminal embedding $\mathbf e_{x}$, or a closed constituent embedding $\mathbf e_{c}$ obtained by REDUCE action.

At each step, the number of candidate actions is $|\mathcal{N}|+2$ given the set of nonterminal symbols $\mathcal{N}$.
Each action changes $S_h$ and $S_e$ as follows:
\begin{itemize}
 \item NT($x$): Push open nonterminal embedding $\mathbf e_x$ onto $S_e$, getting a new LSTM state by $\mathbf h_{\textrm{new}} = \textrm{LSTM}(\textrm{top}(S_h), \mathbf e_x)$, and then push $\mathbf h_{\textrm{new}}$ onto $S_h$.
       This action corresponds to generating an open nonterminal, e.g, ``(VP'' (when $x$=VP), which will be closed later.
 \item GEN: First, generate a next token by sampling from $w \sim \textrm{softmax}(\mathbf W_w \textrm{MLP}(\mathbf h_t) + \mathbf b_w)$.
       Then, as in NT, push $\mathbf e_w$ onto $S_e$, getting a new LSTM state $\mathbf h_{\textrm{new}} = \textrm{LSTM}(\textrm{top}(S_h), \mathbf e_w)$, and push $\mathbf h_{\textrm{new}}$ onto $S_h$.
 \item REDUCE: First, repeatedly pop from $S_e$ $k$-times until we find $\mathbf e_x$, an open nonterminal embedding.
       Letting $\mathbf e^{t-k} = \mathbf e_{x}$, then, apply a composition function, which is BiLSTM~\cite{dyer-etal-2016-recurrent} by default, to obtain a composed phrase representation $\mathbf e_c$:
       \begin{align*}
        \mathbf{e}_c = \textrm{BiLSTM}([\mathbf{e}^{t-k}, \cdots, \mathbf{e}^t]).
       \end{align*}
       $\mathbf{e}_c$ is then pushed onto $S_e$.
       To synchronize two stacks, we also pop $k$-times from $S_h$ and update the LSTM state by $\mathbf h_{\textrm{new}} = \textrm{LSTM}(\textrm{top}(S_h), \mathbf e_c)$, pushing it onto $S_h$.
\end{itemize}

By declaring the operations as above, we notice that the main reasons to prevent mini-batching are twofold:
(1) the stacks have variable length, which varies at each step for each sentence; and more crucially,
(2) internal operations in an action, especially in REDUCE and others, are largely different.

As we describe next, the issue regarding (1) has been largely solved in previous work.
For (2), our strategy is essentially not joining different action types, but trying to improve the efficiency of each action as much as possible after grouping by action types.
We find that in practice this strategy works quite well (Section~\secref{effectsofbatchsize}), allowing models to benefit from a large batch size effectively.

\subsection{Batched stack LSTMs}
\seclabel{batchedstacklstms}
\citet{ding-koehn-2019-parallelizable} propose a sentence-level batched training algorithm for a restricted class of stack LSTMs designed for unlabeled dependency parsing without composition operations~\cite{dyer-etal-2015-transition}.
More specifically, \citet{ding-koehn-2019-parallelizable} deal with the parsing models defined by the following two operations only:\footnote{
A restricted model of an arc-eager system~\cite{nivre-2004-incrementality}, which just POPs when LEFT-ARC occurs, can be achieved with these operations.
RIGHT-ARC is modeled by PUSH.
Essentially, this stack LSTM can only models the right spine of a tree at each step.}
\begin{itemize}
 \item PUSH: Push $\textrm{LSTM}(\textrm{top}(S_h), \mathbf{e}_w)$ to $S_h$.
       $\mathbf{e}_w$ is the embedding of the next token.
 \item POP: Pop the top element from $S_h$.
\end{itemize}
At each step, the next action is either PUSH or POP for each sentence.
This model still suffers from the problem (1) above.
However, they show that by changing the data structure of stack, next PUSH and POP across sentences can be performed in batch.
Given $B$ sentences in a batch, let $S^i_h$ be a stack for $i$-th sentence.
What we need to do is to access all top elements of $S^i_h (i \in [0,\cdots,B-1])$ jointly, and this is possible by summarizing all stacks into a single stack \textit{tensor}, denoted by $\mathbf S_h$, for which $\mathbf S_h[i,p]$ denotes $p$-th element (LSTM state) on the stack for $i$-th sentence.

The core idea behind achieving PUSH and POP jointly is that we perform LSTM updates for \textit{all} stack top elements in a batch, but \textit{only} proceed top stack pointers for PUSH batches.
Given next actions $\mathbf a = $[PUSH, PUSH, POP, $\cdots$] of length $B$, we get a vector $\mathbf {op}=[+1, +1, -1, \cdots]$, denoting whether next stack pointer is +1 (PUSH) or -1 (POP).
By keeping stack top pointer vector $\mathbf p_h$, each step can be batched as the following two operations:
\begin{align*}
 \hspace{-8pt}\mathbf S_h[(0, \mathbf p_h[0] &+ 1) \cdots (B-1, \mathbf p_h[B-1]+1)] \leftarrow \\
 \textrm{LSTM}(\mathbf S_h[(&0, \mathbf p_h[0]) \cdots (B-1, \mathbf p_h[B-1])], \mathbf E_w), \\
 &\mathbf p_h \leftarrow \mathbf p_h + \mathbf{op},
\end{align*}
in which $\mathbf E_w$ is the next token embeddings.

Unfortunately, this batching relies on a strong assumption about models that one action (PUSH) involves all operations (LSTM update and pointer move by $\mathbf{op}$).
This is not the case for RNNGs, for which any action cannot be reduced to a subset of other actions, necessitating a different strategy.

\section{Batched RNNGs}
\seclabel{proposed}
Our batching algorithm for RNNGs is built on the following two observations:
\begin{enumerate}[(a)]
 \item For all $a_t \in \{\textrm{NT}, \textrm{GEN}, \textrm{REDUCE}\}$, the last step is common and corresponds to PUSH operation for stack LSTM above with newly created embeddings $\{\mathbf e_x, \mathbf e_w, \mathbf e_c\}$.
       This final step can be batched if we get all new embeddings as a single tensor $\mathbf{E}_\textrm{next}$ (with size of ($B, |\mathbf e|$)).\label{item:obs1}
 \item Then, the main problem is reduced to getting $\mathbf{E}_\textrm{next}$ efficiently.
       This is possible by separately filling $\mathbf{E}_\textrm{next}$ for each action, using a few additional pointer vectors to keep track of a stack state for each sentence.
\end{enumerate}

To obtain $\mathbf{E}_\textrm{next}$, for NT and GEN, we just need to lookup embeddings for next words and nonterminal symbols.
We need an additional effort to obtain multiple $\mathbf e_c$s at once.
Assuming a stack \textit{tensor} as in~\citet{ding-koehn-2019-parallelizable}, we wish to pop $k$ elements, up to $\mathbf e_c$, for multiple stacks by a single operation.
Now the stack top positions can be accessed by $\mathbf p_h$ (Section~\secref{batchedstacklstms}), which will be the end indices.
To obtain the last open nonterminal positions across a batch, just keeping the last nonterminal positions is insufficient because there are multiple open nonterminals in general.
The following matrix and vector allow this operation:
\begin{itemize}
 \item $\mathbf q$: A matrix of size ($B, D$) given a predefined stack depth bound $D$.
       $\mathbf q[b,d]$ denotes the position of $d$-th nonterminal on the $b$-th stack.
 \item $\mathbf p_q$: A $B$-dimensional vector, pointing to the last index of $\mathbf q$ (similar to $\mathbf p_h$ for $\mathbf S_h$).
\end{itemize}
For example, by $\mathbf q[(0,\mathbf p_q[0])\cdots (B-1,\mathbf p_q[B-1])]$, we can retrieve all the top open nonterminal positions in a batch.
Note that for each $\mathbf q[b]$, the index beyond $\mathbf p_q[b]$ will not be accessed, so we can signify the remove of top nonterminals just by a decrement of $\mathbf p_q$ without updating $\mathbf q$.

\begin{algorithm}[t]
 \small
 \caption{One training step for batched RNNG}\label{algorithm}
 \hspace*{\algorithmicindent} \textbf{Input} Next action vector at $\mathbf{a}$;\\
 \hspace*{\algorithmicindent}\hspace{25pt}index vector for each action: $\mathbf i_{\textrm{gen}}, \mathbf i_{\textrm{nt}}, \mathbf i_{\textrm{red}}$
 \begin{algorithmic}[1]
 \State{$\mathbf E_{\textrm{next}} \leftarrow$ new tensor of size ($B, |\mathbf e_x|$)}
 \State{\color{red}{$\mathbf E_{\textrm{next}}[\mathbf i_\textrm{gen}] \leftarrow$ word\_emb($\mathbf x[(\mathbf i_\textrm{gen}, \mathbf b[\mathbf i_\textrm{gen}])]$)}}
 \State{$\mathbf b[\mathbf i_\textrm{gen}]$ = $\mathbf b[\mathbf i_\textrm{gen}] + 1$} \Comment{Move to next word.} \label{algorithmupdateb}
 \State{\color{red}{$\mathbf E_{\textrm{next}}[\mathbf i_\textrm{nt}] \leftarrow$ nt\_emb($\mathbf a[\mathbf i_\textrm{nt}]$)}}
 \State{$\mathbf p_q[\mathbf i_\textrm{nt}] \leftarrow \mathbf p_q[\mathbf i_\textrm{nt}] + 1$} \label{algorithmupdatepq}
 \State{$\mathbf q[(\mathbf i_\textrm{nt}, \mathbf p_q[\mathbf i_\textrm{nt}])] \leftarrow \mathbf p_h[\mathbf i_\textrm{nt}] + 1$} \Comment{Keep new NT depth.} \label{algorithmupdateq}
 \State{$\mathbf p_{\textrm{prev\_nt}} \leftarrow \mathbf q[(\mathbf i_{\textrm{red}}, \mathbf p_q[\mathbf i_\textrm{red}])]$ }
 \State{$\mathbf E_{\textrm{red}} \leftarrow$ gather\_children($\mathbf p_{\textrm{prev\_nt}}, \mathbf p_h[\mathbf i_\textrm{red}], \mathbf S_{e}$)}
 \State{\color{red}{$\mathbf E_{\textrm{next}}[\mathbf i_\textrm{red}] \leftarrow$ BiLSTM($\mathbf E_{\textrm{red}}$)}} \Comment{Composition.}
 \State{$\mathbf p_q[\mathbf i_\textrm{red}] \leftarrow \mathbf p_q[\mathbf i_\textrm{red}] - 1$} \Comment{Forget about reduced nts.} \label{algorithmupdatepqred}
 \State{$\mathbf p_h[\mathbf i_\textrm{red}] \leftarrow \mathbf p_{\textrm{prev\_nt}} - 1$}
 \State{$\mathbf p_h \leftarrow \mathbf p_h + 1$} \label{algorithmupdatephall}
 \State{$\mathbf S_h[\mathbf p_h] \leftarrow $ LSTM($\mathbf S_h[\mathbf p_h - 1], \mathbf{E}_{\textrm{next}}$)}\label{algorithmshupdate}
 \State{$\mathbf S_e[\mathbf p_h] \leftarrow \mathbf{E}_{\textrm{next}}$}\label{algorithmseupdate}
 \end{algorithmic}
\end{algorithm}

\begin{figure}[t]
 \centering
 \begin{minipage}[t]{\linewidth}
  \centering
  \scalebox{0.8}{
  \begin{tabular}[t]{llllll|l}
   \multicolumn{6}{c|}{Stack}&Next action\\ \hline
   $|_0$ & (S $|_1$   & So $|_2$   & (NP       & it      & ) $|_3$  & (VP  \\
   $|_0$ & (S $|_1$   & (NP        & he        & )       $|_2$  & (VP $|_3$  & said \\
   $|_0$ & (S $|_1$   & (NP $|_2$  & (NP $|_3$ & A $|_4$ & branch $|_5$ & )
  \end{tabular}
  }
 \end{minipage}
 \begin{minipage}[t]{0.4\linewidth}
  {\small
  \vspace{-10pt}\begin{align*}
   \mathbf b &= [1, 1, 2] \\
   \mathbf p_h &= [3, 3, 5] \\
   \mathbf p_q &= [0, 1, 2]
  \end{align*}
  }
 \end{minipage}
 \begin{minipage}[t]{0.4\linewidth}
  {\small
  \vspace{-10pt}\begin{align*}
   \mathbf q = \left[
   \begin{array}{rrrr}
    \underline{1}& 3& 0& \cdots\\
    1& \underline{3}& 0& \cdots\\
    1& 2& \underline{3}& \cdots\\
   \end{array}
   \right]
  \end{align*}
  }
 \end{minipage}
 \vspace{-20pt}\caption{Example batched stack configuration.
 $x |_d$ means that $x$ is at depth $d$.
 For example, ``(NP it)'' in the first sentence is closed so constitutes a single item on the stack.
 $\mathbf p_q$ points to the top positions of $\mathbf q$, which are underlined.
 }\figlabel{batchedstack}
\end{figure}

We need a few additional tensors to achieve fully batched stack tensor operations.
Figure~\figref{batchedstack} shows an example.
\begin{itemize}
 \item $\mathbf S_h$: A tensor of size ($B, D, L, H$) when the stack LSTM has $L$ layers with $H$ hidden dimensions.
       The core of batched stack LSTMs.
 \item $\mathbf S_e$: A tensor of size ($B, D, |\mathbf e|$), corresponding to $S_e$ in non-batched models~(Section \secref{rnng}).
 \item $\mathbf b$: A $B$-dimensional vector, keeping the next token index in the sentence.
 \item $\mathbf p_h$: A $B$-dimensional vector, pointing to the top elements of $\mathbf S_h$.
\end{itemize}

\paragraph{Full algorithm}
Algorithm~\ref{algorithm} describes operations in each step given next actions $\mathbf a$.
Action index vector $\mathbf i_a$, keeps the indices of action $a$ in $\mathbf a$;
in Figure~\figref{batchedstack}, $\mathbf i_\textrm{gen} = [1]$ and $\mathbf i_\textrm{nt} = [0]$.
The operations are mainly categorized into filling $\mathbf E_{\textrm{next}}$ for each action (in red), pointer updates according to action definitions (\ref{algorithmupdateb}, \ref{algorithmupdatepq}, \ref{algorithmupdateq}, \ref{algorithmupdatepqred}, \ref{algorithmupdatephall}), and finally stack updates (\ref{algorithmshupdate}, \ref{algorithmseupdate}), corresponding to the observed common operations (\ref{item:obs1}).\footnote{
By $A[(\mathbf x, \mathbf y)]$ for vectors $\mathbf x$ and $\mathbf y$, both with length $l$, we mean $A[(x[0], y[0]) \cdots (x[l-1], y[l-1])]$, corresponding to advanced indexing in PyTorch.
We regard that $\mathbf S_h[0]$ is fixed by initial hidden vectors while $\mathbf S_e[0]$ is kept empty.
$\mathbf p_h[b] = 0$ means that $b$-th stack is empty.
}
\textit{gather\_children} is a function that returns a tensor summarizing reduced children node embeddings.
Since the number of reduced children differs across batch, we implement this to return a padded tensor, using \textit{gather} function in PyTorch.

Deviated from~\citet{ding-koehn-2019-parallelizable}, we \textit{separately} perform each action, as indicated by the use of $\mathbf i_a$.
This can be seen as a deficiency of our algorithm;
however, this separation is necessary beyond very simple models, which are practically less attractive.
Rather, our strategy can be applied to broader classes of structured neural models, including dependency parsing with composition, and we believe that our empirical success (Section~\secref{expbllip}) encourages further exploration of the presented strategy to various models.

\paragraph{How to set $D$?}
As in~\citet{ding-koehn-2019-parallelizable}, we need to specify the stack depth bound $D$ for each batch.
Increasing this value incurs more GPU memory.
For training, we can precompute the minimum value of each sentence by simulating oracle transitions beforehand and use it.
For inference, we fix $D=100$, since we find that even for very long sentences (more than 150 words), the stack depth will never exceed 80 for English sentences.

\paragraph{A note on extra memory with stack tensors}
At first sight, our approach seems to suffer from the limitation in scalability due to fixed stack tensors ($\mathbf S_h$ and $\mathbf S_e$).
The sizes of these tensors grow by model size, implying that we may not be able to employ a large batch size for a large model.
In practice, however, this extra memory will not be a bottleneck in the total memory for training.
This is because the main cause of required memory during training is rather a computational graph itself, which keeps all intermediate hidden states at each step.
Our stack tensors can be seen as a ``storage'' to allow computing these intermediate values effectively with tensor operations.
The extra memory for this storage is smaller than the total memory in a computational graph because the former depends on $D$ while the latter depends on the total action length $A$, and $D \ll A$  in general.\footnote{
Our preliminary experiment suggests that our RNNG implementation can be scaled at least comparable model and data sizes to ELMo~\cite{peters-etal-2018-deep}, a large-scale LSTM-based model, given a similar amount of computing resources.
We examine the maximum allowable batch size for a model with 1,256 hidden dimensions, amounting to 94M parameters, which are comparable to ELMo (93M), and find that the batch size can be increased to 256, with the maximum action size in a batch of 16,000 (see Section~\secref{speedsetting}) on a single V100 GPU (16GB).
Transformer-level scalability~\cite{devlin-etal-2019-bert} would still be infeasible because of the RNNG's limited paralellism that is only on sentence-level, not token-level as in Transformers.
}

\section{Other Improvements}
\seclabel{otherimprovements}
\paragraph{Batched beam search}
For inference as a language model or as an incremental parser, RNNGs typically employ word-synchronous beam search~\cite{stern-etal-2017-effective,hale-etal-2018-finding}, which is although known to be very slow~\cite{crabbe-etal-2019-variable} because it often requires large beam sizes, such as 100 or 1000, and operations are not batched.
As a by-product of our batched training, we succeed at implementing fully batched beam search for RNNGs, excluding any \textit{for} loops, by which we drastically improve the search speed (Section~\secref{beamsearchspeed}).
This is possible by adding the ``beam'' dimension to all state tensors ($\mathbf S_h$, $\mathbf q$, etc.).

\paragraph{Subwords}
Given an increased amount of training data, the vocabulary size naturally increases.
To suppress this effect, using subwords~\cite{sennrich-etal-2016-neural} now becomes a standard technique.
We thus incorporate subword modeling into our RNNGs and employ it for our largest experiment in Section~\secref{syntaxgym}.
\citet{doi:10.1162/tacl-a-00345} recently incorporate subwords in RNNGs, in which, each word is regarded as a new constituent with WORD label, e.g., (WORD cu$|$ r$|$ ry).
This means that models always need to perform additional NT(WORD) and REDUCE for each \textit{token}, even for unsegmented ones, e.g., (WORD I), greatly increasing the average action sequence length, which in turn affect the training time.
In this work, we model subwords by a simpler method of just segmenting each token.
For example, an NP looks like (NP Th$|$ ai cu$|$ r$|$ ry).
While \citet{doi:10.1162/tacl-a-00345} note that this simple modeling is less effective, our experiments suggest that this is a good enough strategy, considering the added computational costs with NT(WORD) actions.\footnote{
We provide a pilot study about this method in Appendix~\secref{expsubwords}.
Using BLLIP corpus and Penn Treebank, we explore the relationship between a suitable number of subword units and model sizes.
The main result is that large subword units are effective for larger models, and also subword modeling almost always improves parsing accuracy.
}

\section{Evaluating Efficiency of Batching}
\seclabel{expbllip}
The main focus of this section is a comparative evaluation of our PyTorch RNNG implementation with the existing DyNet implementation.\footnote{
https://github.com/cpllab/rnng-incremental.
This implementation supports word-synchronous beam search.
For the training part, this is not implemented to use DyNet Autobatch so we modified it to enable that.
}
We show that: (1) with a large batch size training speed drastically improves, and models will tend to find better parameters (Section~\secref{effectsofbatchsize}); and
(2) our batched beam search hugely speeds up inference (Section~\secref{beamsearchspeed}).

\subsection{Setting}
\seclabel{speedsetting}
While Penn Treebank (PTB;~\citealp{marcus-etal-1993-building}) has often been used to train RNNGs~\cite{wilcox-etal-2019-structural,wilcox-etal-2020-structural}, it is too small and here we use a larger dataset of BLLIP corpus~\cite{bllip}, expecting that the effects of large batch size become clearer by this modestly sized dataset.

\paragraph{Preprocessing}
For preprocessing, we largely follow \citet{hu-etal-2020-systematic}, which also train an RNNG on this dataset.
We partition the data according to their \textsc{lg} size, the largest training setting, amounting to 42 million tokens for training and 1,500 sentences for development.
One difference we make is the handling of unknown tokens.
We limit the vocabulary by the top frequent 50,000 word types in the training data.
\citet{hu-etal-2020-systematic} use all word types that appear at least twice; 
however, this method vastly increases the vocabulary size and hence the model size.
Unknown tokens are created in the same way with the Berkeley parser's surface feature rule~\cite{petrov-etal-2006-learning}.
The way to annotate parse trees is the same as well;
we run Berkeley neural parser~\cite{kitaev-etal-2019-multilingual}, a state-of-the-art constituency parser to assign accurate parses.

\paragraph{Model size and parameters}
We experiment with the most common model size of RNNG in the literature: 256 dimensions for input and LSTM hidden dimensions, with 2 layer LSTMs~\cite{dyer-etal-2016-recurrent,hu-etal-2020-systematic}.
The total number of parameters is about 15M.
The hyperparameters are summarized in Appendix~\secref{hyper}.

\begin{figure}[t]
 \centering
 \includegraphics[width=0.85\linewidth]{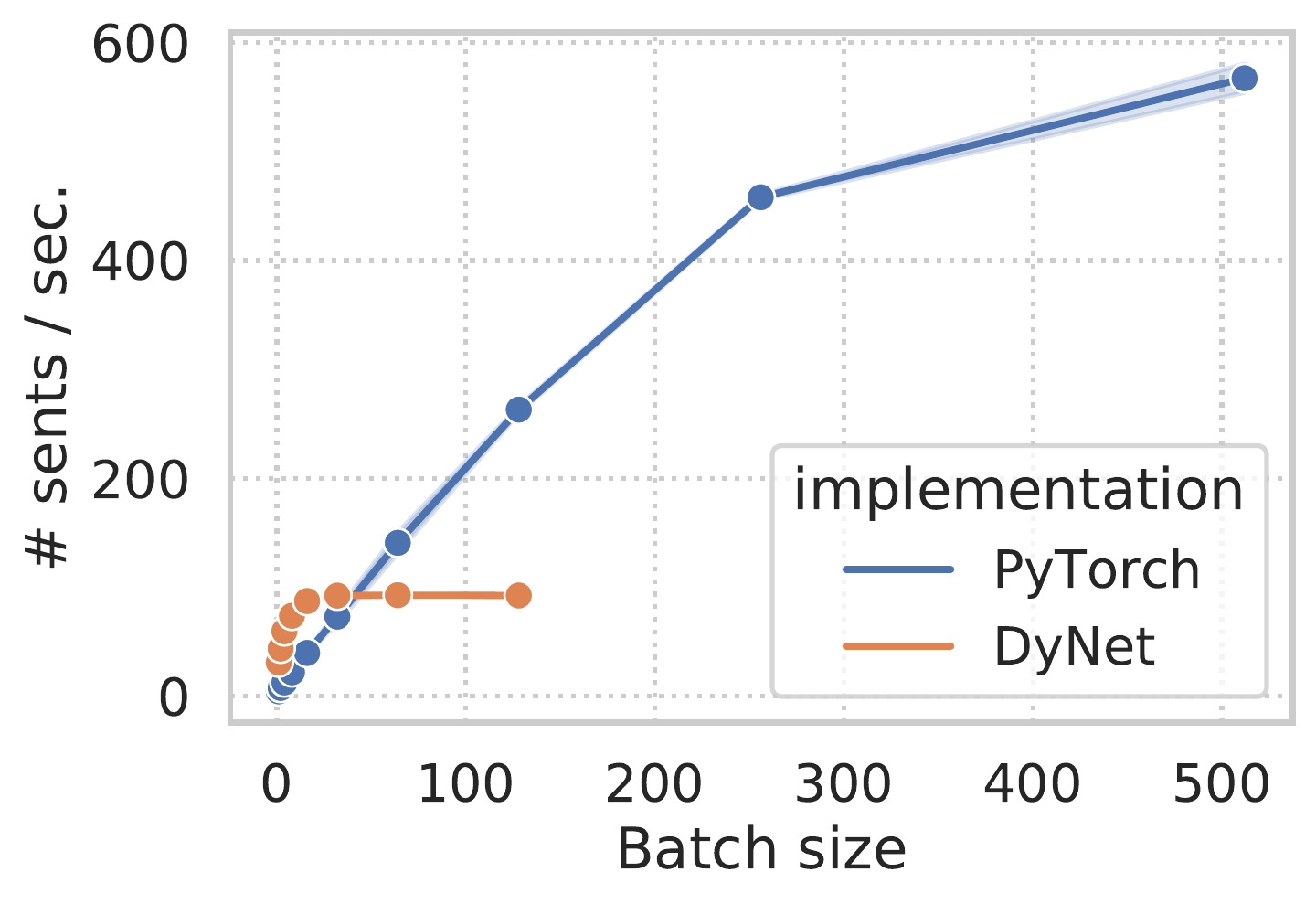}
  \caption{Training speed comparison (number of processed sentences / sec.) when increasing batch sizes to [1, 2, 4, $\cdots$, 512].
 Shade denotes standard deviation.}\figlabel{pytorchdynetspeed}
\end{figure}

\paragraph{Other settings}
We employ some additional techniques to improve the efficiency of our batching mechanism.
First, before training, we group sentences by their number of gold actions so that examples in each mini-batch have similar numbers of actions.
Specifically, we first sort the sentences by action lengths, divide by every 4096 sentences, and then sample each batch from a single group.

Second, we predefine the maximum value for the total number of actions across sentences in a batch, which we set to 26,000.
This is inspired by a similar mechanism in fairseq~\cite{ott-etal-2019-fairseq} for the maximum number of tokens.
Using this means that the number of sentences in a batch will be adjusted to be smaller than the batch size when the action sequences (or sentences) are long, allowing us to interpret given batch size as the maximum that is fully exploited only for shorter sentences, which are in practice dominant in the data.\footnote{
To reduce the memory further, we also employ mixed-precision training in PyTorch.
Inference is performed with half-precision (fp16).
We find that this does not change the results of beam search at all.
}

Every experiment is run on a single V100 GPU with 16GB memory.
Unless otherwise noted, we perform every experiment three times with different random seeds, reporting an average score with standard deviation.

\begin{figure}[t]
 \centering
 \includegraphics[width=0.85\linewidth]{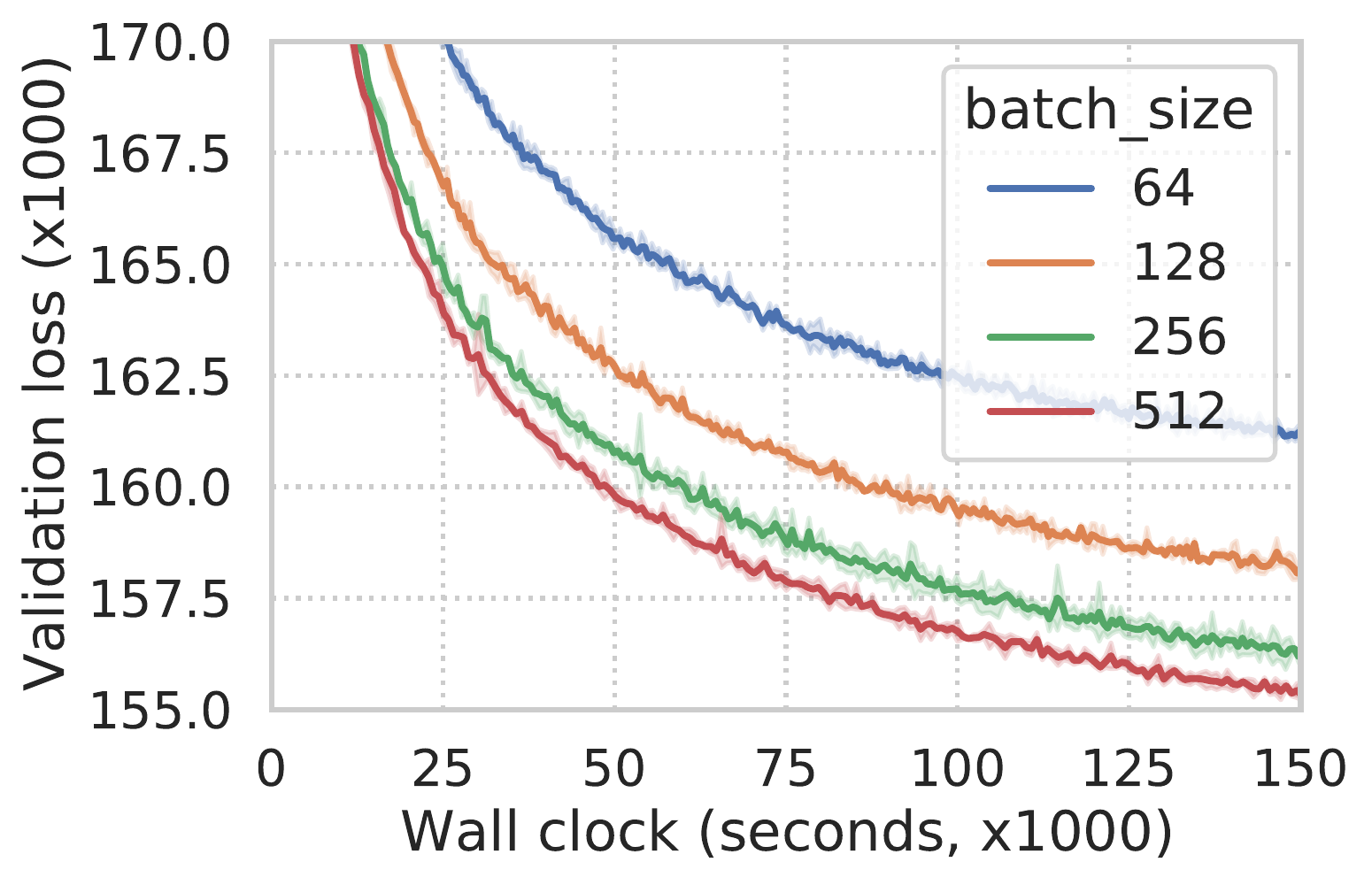}
 \caption{Training wall clock time vs.~total validation loss for different batch sizes.
 X-value of $i$-th point is an averaged duration time to $i$-th validation step across three random seeds.
 Shade denotes standard deviation.
 }\figlabel{batchvsloss}
\end{figure}

\subsection{Effects of batch sizes}
\seclabel{effectsofbatchsize}
Although our batched training involves action-specific operations (Section~\secref{proposed}), to our surprise, the efficiency improvement for our RNNG with large batch sizes is almost linear up to 256 (Figure~\figref{pytorchdynetspeed}).
The improvement is narrow at 512, though this is mainly due to the restriction of the maximum number of actions in a batch (Section~\secref{speedsetting}), which reduces the actual batch size for longer inputs.
DyNet's Autobatch is quite effective up to 16, running much faster than ours due to the speed of C++, but further improvement is not obtained probably because of the increased overhead of finding mini-batch units themselves from a large computational graph.

Though this result clearly demonstrates the efficiency of our batching mechanism, it is \textit{only} meaningful when the large batch size in fact leads to a faster model \textit{convergence}.
This is the case, as shown in Figure~\figref{batchvsloss}, where we compare the total validation losses as a function of actual wall clock time during training.
The loss is calculated every 1000 batches.
The model with batch size 512 converges fastest, and importantly, to better parameters.
This result suggests that we can safely benefit from large batch size as long as memory permits.
In the following experiments, we fix the batch size to 512.

\subsection{Beam search speed improvement}
\seclabel{beamsearchspeed}

As we discuss in Section~\secref{otherimprovements}, we have also improved the efficiency of word-synchronous beam search, a standard technique to calculate incremental prefix probabilities~\cite{hale-2001-probabilistic} and a parse tree for RNNGs.
Now, we evaluate the impact of this improvement.
For PyTorch, we run it on V100 GPU;
for DyNet, we find that it runs faster on CPUs so we instead use CPUs (Intel Xeon 6148, 20 cores x2), with Intel MKL.
DyNet beam search is still too slow with this environment so we limit the number of tested sentences to 300 from the BLLIP development set.
For PyTorch, we try two different batch sizes \{1, 10\}, with a restriction on the number of tokens in a batch, similarly to the total action size in training (Section~\secref{speedsetting}).
We fix this value to 250, with which the model can safely parse with the largest beam size of 1000.

Word-synchronous beam search employs two types of beam widths, action beam size ($k$) and word beam size ($k_w$), along with fast-tracked candidate size, denoted as $k_s$ (see~\citealt{stern-etal-2017-effective}).
$k$ is most akin to the standard beam size.
Table~\tabref{beamsearchspeed} summarizes the results when increasing $k$ (others are in the caption).
The beam search of DyNet becomes prohibitively slow when $k \geq$ 50.
Strikingly, the increase in average runtime is more than linear against the beam size, especially for 10$\rightarrow$50 and 50$\rightarrow$100.
The time increases, 0.5$\rightarrow$11.3 (x22.6) and 11.3$\rightarrow$48.6 (x4.3) are roughly quadratic to the increase of $k$ (x5 and x2).
This result is reasonable because in addition to the complexity of each step, which depends on $k$,
the length of searched action sequence could also linearly grow by $k$.\footnote{
For a sentence of length $N$, the runtime of beam search is $O(k \times N \times M_w)$, where $M_w$ denotes the maximum number of actions between two tokens (until choosing next SHIFT).
The expected number of actions between two tokens (bound by $M_w$) grows by $k$ because at each step, with a large $k$ the chance that non-shift beam items remain in the next beam increases;
hence, the runtime becomes quadratic to $k$ in the worst case.
We conjecture that this inefficiency is bounded at some $k$ (see $k=200$), though is severe for smaller $k$s.
}
The naïve DyNet implementation directly suffers from this computational cost.

\begin{table}
\centering
\scalebox{0.73}{
\begin{tabular}{lrrrrrr}
\toprule
Action beam size $k$ & \multicolumn{1}{c}{10}&\multicolumn{1}{c}{50}&\multicolumn{1}{c}{100}&\multicolumn{1}{c}{200}&\multicolumn{1}{c}{400}&\multicolumn{1}{c}{1000} \\
\midrule
DyNet&0.5&11.3&48.6&100.6&201.3&NA\\
PyTorch (B=1)&1.7&2.1&2.3&2.5&2.9&4.1\\
PyTorch (B=10)&0.4&0.5&0.7&0.9&1.3&2.8\\
\bottomrule
\end{tabular}
}
 \caption{Word-synchronous beam search speed (average seconds per sentence) comparison on the first 300 sentences in BLLIP development set.
 B denotes the batch size.
 Word beam sizes ($k_w$) / fast track sizes ($k_s$) are 10/1, 10/1, 10/1, 20/2, 40/4, and 100/10, respectively.
 }\tablabel{beamsearchspeed}
\end{table}

Our batched beam search largely resolves this issue and now the average runtime only gradually increases by $k$.
We note that as a parser or a language model, this speed is still not very fast, considering that this is on a GPU.\footnote{
For smaller $k$, we can increase the batch size and the maximum number of tokens in a batch to further speedup.
}
For the research purpose, including psycholinguistic assessments as done in Section~\secref{syntaxgym}, however, this improvement is significant, making experiments much easier even with large beam sizes.
We still need to work on improving efficiency further, possibly by modifying learning methods to replace word-synchronous search~\cite{stanojevic-steedman-2020-max}.

\section{Syntactic Generalization Performance of Scaled RNNG}
\seclabel{syntaxgym}

\begin{figure*}[t]
 \centering
 \includegraphics[width=0.95\linewidth]{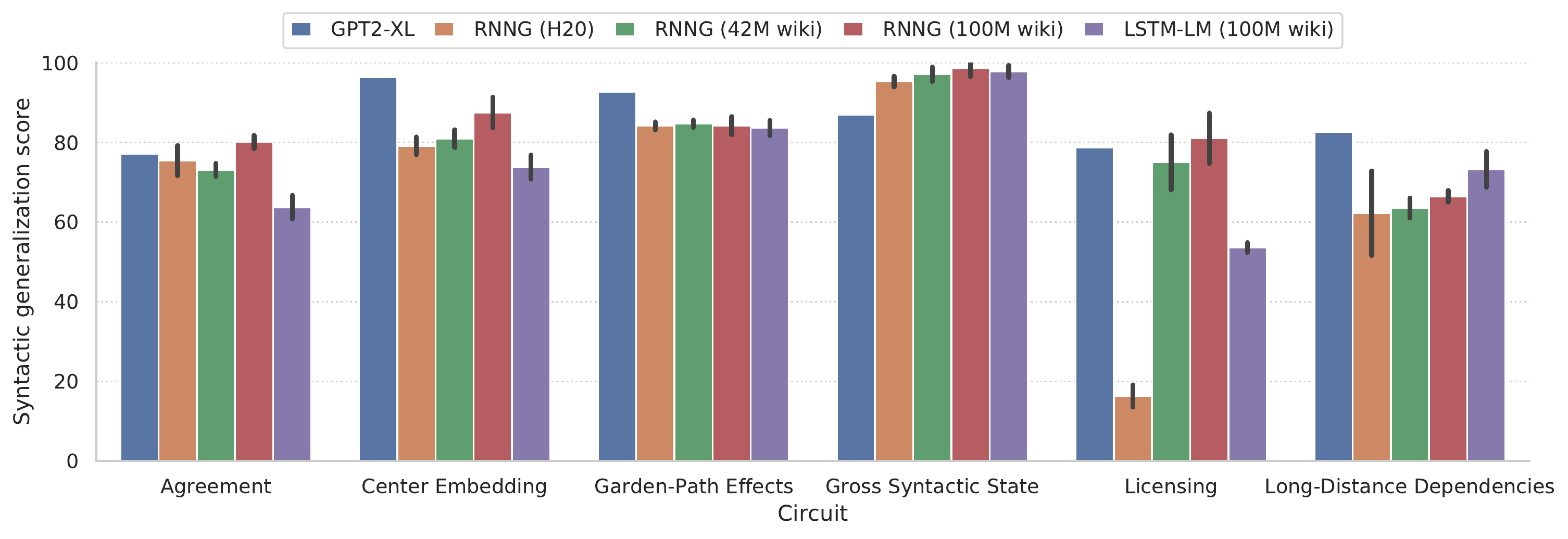}
 \caption{Circuit-level accuracies on SyntaxGym. For each circuit, suite-level accuracies are averaged across different test suites and random seeds to compute ``Syntactic generalization score'' of each model. Note that RNNG (H20) is a model trained on BLLIP (about 40M tokens) in \citet{hu-etal-2020-systematic} but diverged from their results, because their suite-level accuracies are averaged across different models trained on various data sizes.
 }\figlabel{circuitaccuracy}
\end{figure*}

Finally, we evaluate the syntactic generalization abilities of the scaled RNNG.
For this purpose, we adopt the test circuits used in~\citet{hu-etal-2020-systematic} via SyntaxGym~\cite{gauthier-etal-2020-syntaxgym}.
Here, a test \textit{circuit} is a collection of test \textit{suites};
e.g., ``Long-Distance Dependencies'' circuit contains a suite on a specific type of ``filler-gap dependencies'' as well as a suite on (pseudo) ``cleft''.
For each example in a suite, a model succeeds if it can assign a higher likelihood on a grammatically critical position in the correct sentence.
For example, given ``\textit{The farmer near the clerks \underline{knows/*know} many people.}'' in the ``Agreement'' circuit, a model is correct if it assigns $p(\textit{knows}|h) > p(\textit{know}|h)$.
Note that for subword models the total likelihoods on subwords (not averaged) are compared.

In the previous literature, \citet{hu-etal-2020-systematic} trained an RNNG on BLLIP (42M tokens).
Here, we train subword RNNGs on 100M tokens from English Wikipedia, to which we assign parse trees with Berkeley neural parser.
The model size is 35M with 30k subword units, following the experiment in Appendix~\secref{expsubwords}, which assesses the suitable number of subword units for different model sizes.
We train this RNNG for three days (with three different seeds), and at inference fix the beam size $k$ to 100 ($k_w=10, k_s=1$).
We also train an RNNG with the subset of this data (42M tokens) to separate the effects of data size.
Our LSTM baseline is the one used in \citet{noji-takamura-2020-analysis}, which is basically an AWD-LSTM-LM~\cite{merity2018regularizing} extended to be sentence-level and, on the~\citet{marvin-linzen-2018-targeted} benchmark, shown to work better than GRNN~\cite{gulordava-etal-2018-colorless}, one of the models used in~\citet{hu-etal-2020-systematic}.\footnote{Our LSTM implementation is available at https://github.com/aistairc/lm\_syntax\_negative. We train this LSTM on our subword-segmented Wikipedia (30k units). The model size is adjusted so that the total number of parameters becomes 35M, the same size as our RNNGs (3 layer LSTMs with 1150 hidden and 450 input dimensions).}

The main result on circuit-level accuracies is summarized in Figure~\figref{circuitaccuracy}.
On the effects of the data scale, we observe a consistent improvement from ``RNNG (42M wiki)`` to ``RNNG (100M wiki)''.
This result suggests that this amount of increase in training data is still beneficial for structural language models to strengthen their syntactic generalization ability.
For some circuits, only RNNG (100M) outperforms GPT-2~\cite{radford2019language} on average (``Agreement'' and ``Licensing'').

Comparing LSTM (100M) and RNNG (100M), RNNG generally outperforms LSTM, but with an exception on ``Long-Distance Dependencies''.
In order to inspect this, we break down this circuit into suites, (Figure~\figref{longdistance}), finding that this deficiency of RNNG is due to its poor performance on (pseudo) ``cleft'', including the following example:

\begin{exe}
 \ex
 \begin{xlist}
  \ex \label{cleftcorrect} What he did was \underline{prepare the meal} .
  \ex \label{cleftwrong} *What he ate was \underline{prepare the meal} .
 \end{xlist}
\end{exe}

On underlined tokens, models should assign a higher likelihood for (\ref{cleftcorrect}).
Our LSTM performs nearly perfectly on these cases while our RNNGs perform badly.
We conjecture that given more data and/or parameters, RNNGs will tend to strengthen their commitment to provided syntactic supervision, and hence may lose some lexical heuristics which LSTMs can exploit from surface patterns (e.g., an association of \textit{did} $\rightarrow$ \textit{prepare}).
In fact, the ability of LSTM on cleft is rather brittle, as shown in a huge drop on ``cleft\_modifier'', which include cleft constructions with intervening modifiers.

\begin{figure}[t]
 \centering
 \includegraphics[width=\linewidth]{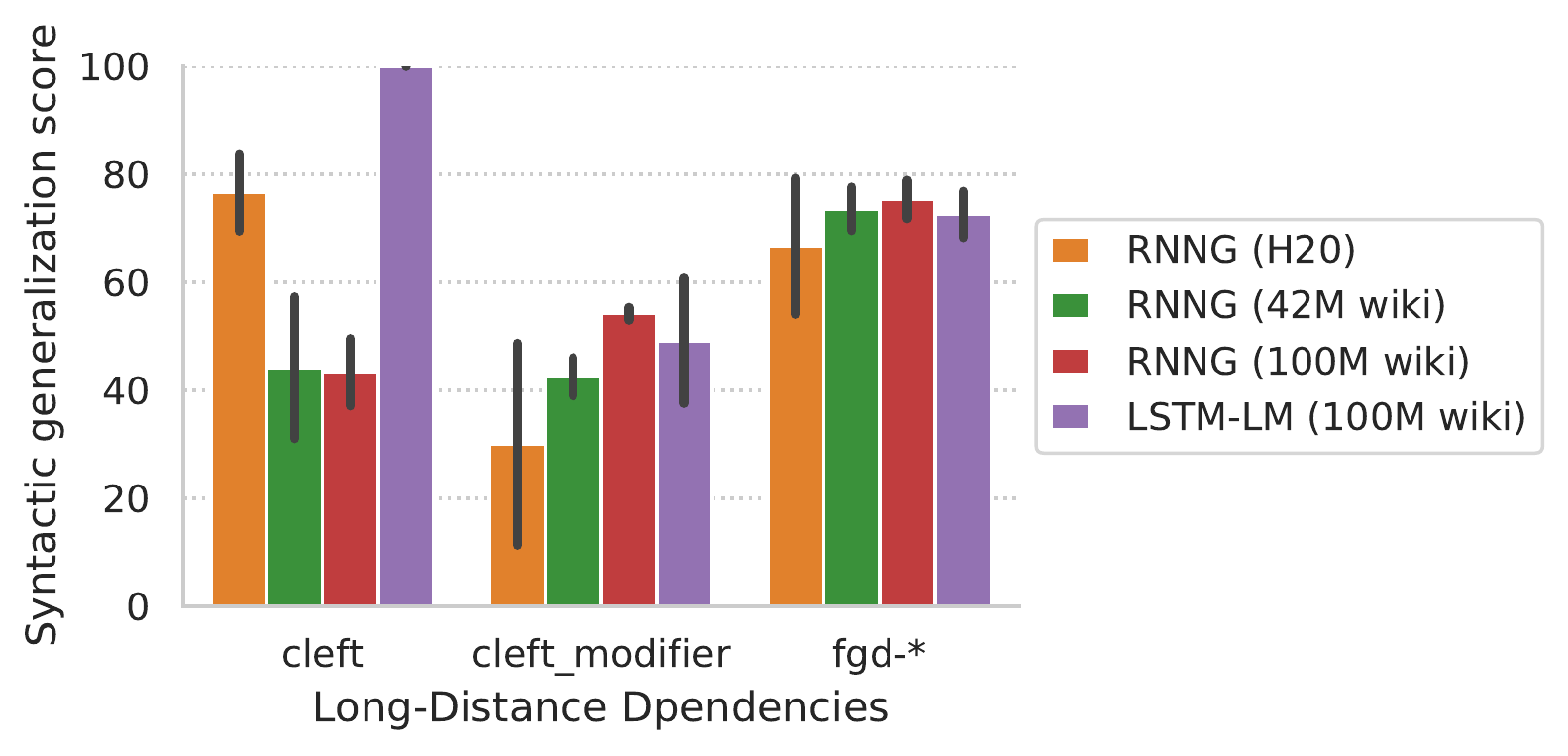}
  \caption{Suite-level accuracies on the ``Long-Distance Dependencies'' circuit. fgd-* is averaged across different test suites of filler-gap dependencies.}\figlabel{longdistance}
\end{figure}

To rigorously handle these cases, models should notice that (\ref{cleftwrong}) is a free relative clause and do not have an antecedent.
However, the currently employed PTB annotation, which is limited to local structures, does not provide a distinction between these clause types, analyzing both as ``(SBAR (WHNP What) (S (NP he) (VP did/ate)))'', which our RNNGs predict correctly.
We also notice that \citet{hu-etal-2020-systematic}'s RNNG (H20) performs rather similarly to our LSTM, while our RNNG (42M), trained on the comparable size of data to H20, is more similar to our RNNG (100M), suggesting that RNNG's poor performance on cleft is not just due to the data scale.
One possible explanation of the discrepancy between H20 and our 42M RNNG is that our RNNG might be better optimized thanks to improved training, or due to the sizes of hidden layers (256 for H20 and 656 for ours).

This problem poses a new interesting challenge.
While RNNGs have been compared to LSTMs several times, the provided syntactic structures are fixed and effects of different annotations (formalism, quantity, etc.) are not explored.
For such investigation, the training cost of RNNGs has been a practical burden, but that problem largely goes away with the current study.
We expect that our new implementation and batching strategy provide fruitful future research opportunities on structured neural language models.

\section{Conclusion}
A large computational cost of training structured neural language models was a main practical burden for employing these models in applications and analyses. With special focus on RNNGs, we have provided a direct solution to this problem by showing that batched effective training is in fact possible. On the large scale experiments with SyntaxGym test circuits, we found that the data quantity further strengthens the syntactic generalization abilities of RNNGs, while the annotation quality or quantity will also be of practical importance towards a language model with human-like strong syntactic performance.

\section*{Acknowledgement}
This paper is based on results obtained from a project, JPNP20006, commissioned by the New Energy and Industrial Technology Development Organization (NEDO). This work was also supported by JSPS KAKENHI Grant Numbers 20K19877 and 19H04990, and the National Institute for Japanese Language and Linguistics (NINJAL) Collaborative Research Project ``Computational Psycholinguistics of Language Processing with Large Corpora.''

\bibliographystyle{acl_natbib}
\bibliography{anthology,acl2021}

\clearpage
\appendix

\section{Hyperparameters}
\seclabel{hyper}
We use the defualt parameter setting for the DyNet implementation.
For our implementation, we use Adam optimizer~\cite{DBLP:journals/corr/KingmaB14}, which is found to be superior, while SGD has been used for DyNet implementation~\cite{dyer-etal-2016-recurrent,wilcox-etal-2019-structural}.
We set the learning rate and dropout rate to 0.001 and 0.1, respectively, which we find achieve lower validation loss robustly across different batch sizes.

\section{Effect of Number of Subword Units}
\seclabel{expsubwords}

We perform an experiment to understand the behaviors of our simple subword modeling (Section~\secref{otherimprovements}).
We use the BLLIP corpus as preprocessed in Section~\secref{speedsetting} except the setting about vocabulary.
We compare the fixed vocabulary models, which we train in the experiment of Section~\secref{effectsofbatchsize} (batch size 512), and several subword vocabulary models.
The hyperparameters are the same as the fixed vocabulary models.

\paragraph{Subword units} The number of subwords can be seen as a hyperparameter.
To understand the effects of this size for RNNGs, we prepare three different subword vocabularies: 10k (10,240), 20k (20,480), and 30k (30,720).
We use byte-pair encoding \cite{sennrich-etal-2016-neural} implemented in sentencepiece \cite{kudo-richardson-2018-sentencepiece}.

\paragraph{Model sizes} We prepare two different model sizes, 15M and 35M, to see the interaction between the suitable size of subword units and model size, by adjusting the number of two dimensions so that the total number of parameters becomes comparable to these numbers.
For 15M parameter models, the dimensions are 528 for 10k units; 432 for 20k; and 336 for 30k.
For 35M parameter models, these are 864, 752, and 656, respectively.
The number of LSTM layers is fixed to 2.

\begin{table}[t]
\centering
\scalebox{0.8}{
\begin{tabular}{lrrrr}
\toprule
$V$ (\# params.) / beam $k$ & \multicolumn{1}{c}{100}&\multicolumn{1}{c}{200}&\multicolumn{1}{c}{400}&\multicolumn{1}{c}{1000} \\
\midrule
$V_{fix}=50k$ (15M)&92.34&93.70&94.26&94.59\\
\midrule
$V_{sb}=10k$ (15M)&92.74&93.85&94.37&94.64\\
$V_{sb}=20k$ (15M)&\textbf{92.77}&\textbf{93.95}&\textbf{94.48}&\textbf{94.80}\\
$V_{sb}=30k$ (15M)&92.33&93.76&94.38&94.72\\
\midrule
$V_{sb}=10k$ (35M)&92.67&93.87&94.37&94.59\\
$V_{sb}=20k$ (35M)&92.84&\textbf{93.93}&\textbf{94.50}&\textbf{94.79}\\
$V_{sb}=30k$ (35M)&\textbf{92.92}&93.84&94.44&94.72\\
\midrule
\citet{hale-etal-2018-finding}&87.1&88.96&90.48&90.96\\
\bottomrule
\end{tabular}
}
 \caption{
 PTB development set parsing accuracy (F1) when changing beam size, averaged on three models with different random seeds.
 $V_{fix}$ is the vocabulary size for fixed vocabulary models while $V_{sb}$ is that for subword models.
 \citet{hale-etal-2018-finding} is trained only on PTB training set and is not directly comparable.
 Word beam $k_{w}=k/10$ and $k_{s}=k/100$.
 }
 \tablabel{subwordparsing}
\end{table}

\seclabel{effectsofsubwords}
\begin{table}[t]
\centering
\scalebox{0.8}{
\begin{tabular}{lrrrr}
\toprule
$V$ (\# params.) / beam $k$ & \multicolumn{1}{c}{100}&\multicolumn{1}{c}{200}&\multicolumn{1}{c}{400}&\multicolumn{1}{c}{1000} \\
\midrule
$V_{fix}=50k$ (15M)&52.34&49.53&48.26&47.53\\
\midrule
$V_{sb}=10k$ (15M)&69.09&65.34&63.74&62.81\\
$V_{sb}=20k$ (15M)&\textbf{67.52}&\textbf{64.15}&\textbf{62.41}&\textbf{61.36}\\
$V_{sb}=30k$ (15M)&70.43&66.41&64.50&63.41\\
\midrule
$V_{sb}=10k$ (35M)&67.66&64.13&62.41&61.47\\
$V_{sb}=20k$ (35M)&63.60&60.33&58.78&57.90\\
$V_{sb}=30k$ (35M)&\textbf{60.80}&\textbf{57.91}&\textbf{56.28}&\textbf{55.45}\\
\bottomrule
\end{tabular}
}
 \caption{Perplexity on BLLIP validation set for each setting described in Table~\tabref{subwordparsing}, averaged on three models with different seeds.
 For $V_{sb}$, perplexity is not subword-level but token-level, by summing subword likelihoods for each token.
 }\tablabel{subwordperplexity}
\end{table}

\paragraph{Results}
We investigate (1) the effectiveness of our simple subword modeling itself, and
 (2) whether the optimal number of subword units depends on model sizes.
For (1), one way of evaluation is to compare the perplexities of subword models and fixed vocabulary models (see Section~\secref{speedsetting}).
However, they are not directly comparable because the fixed vocabulary models replace many tokens with unknown tokens, which are easy to predict and make the comparison unfair~\cite{mielke-etal-2019-kind}.

We instead validate the effectiveness of our subword modeling by not language modeling, but parsing performance.
Note that the text in the BLLIP corpus is Wall Street Journal, the same as Penn Treebank (PTB).
Thus, we expect that the quality of auto parses provided to our training data is high,
allowing us to assume that a good model should parse the gold PTB data more accurately.
We run beam search on the PTB development (section 22) for each model and the results are summarized in Table~\tabref{subwordparsing}.
We can see that F1 scores consistently improve by subword modeling compared to the fixed vocabulary setting.
The effects of model size (15M vs.~35M) are negligible, suggesting that the upper bound parsing performance using the current silver quality data can be reached with smaller models.

For (2) above, while comparing perplexities across subword and fixed-vocabulary models is impossible, comparing different subword units is possible by casting the subword-level likelihoods to token-level likelihoods~\cite{Mie2016Can}.
Table~\tabref{subwordperplexity} summarizes those values along with the results by fixed vocabulary models as reference.
The effects of the number of subword units ($V_{sb}$) are clearer.
For 15M models, the optimal $V_{sb}$ is 20$k$, while for larger 35M models, the optimal size is 30$k$.
This suggests that more parameters are needed to obtain better results for large models.

\end{document}